\documentclass[letterpaper]{article} %
\usepackage{aaai23}  %
\usepackage{times}  %
\usepackage{helvet}  %
\usepackage{courier}  %
\usepackage[hyphens]{url}  %
\usepackage{graphicx} %
\usepackage{natbib}  %
\usepackage{caption} %
\usepackage{algorithm}
\usepackage{algorithmic}

\usepackage{newfloat}
\usepackage{listings}
\DeclareCaptionStyle{ruled}{labelfont=normalfont,labelsep=colon,strut=off} %
\floatstyle{ruled}
\newfloat{listing}{tb}{lst}{}
\floatname{listing}{Listing}
\usepackage[utf8]{inputenc} %
\usepackage{booktabs}       %
\usepackage{amsfonts}       %
\usepackage{nicefrac}       %
\usepackage{microtype}      %
\usepackage{xcolor}         %
\usepackage{latexsym}
\usepackage{multirow}
\usepackage{wrapfig}
\usepackage{subcaption}
\usepackage[leftcaption]{sidecap}
\usepackage{amssymb}
\usepackage{amsmath}
\usepackage{cleveref}
\usepackage{adjustbox}

\newcommand{\lev}{\emph{Leviathan Ornaments}}
\newcommand{\areo}{\emph{Areopagitica}}

\title{Contrastive Attention Networks for Attribution of Early Modern Print}
\author {
    Nikolai Vogler\textsuperscript{\rm 1},
    Kartik Goyal\textsuperscript{\rm 2},
    Kishore PV Reddy\textsuperscript{\rm 1},
    Elizaveta Pertseva\textsuperscript{\rm 1},
    Samuel V. Lemley\textsuperscript{\rm 3},
    Christopher N. Warren\textsuperscript{\rm 3},
    Max G'Sell\textsuperscript{\rm 3},
    Taylor Berg-Kirkpatrick\textsuperscript{\rm 1}
}
\affiliations {
    \textsuperscript{\rm 1} University of California, San Diego\\
    \textsuperscript{\rm 2} Toyota Technological Institute at Chicago\\
    \textsuperscript{\rm 3} Carnegie Mellon University\\
    nvogler@ucsd.edu, kartikgo@ttic.edu, cnwarren@cmu.edu, tberg@ucsd.edu
}

\begin{document}

\maketitle

\begin{abstract}
In this paper, we develop machine learning techniques to identify unknown printers in early modern (c.~1500--1800) English printed books.
Specifically, we focus on matching uniquely damaged character type-imprints in anonymously printed books to works with known printers in order to provide evidence of their origins.
Until now, this work has been limited to manual investigations by analytical bibliographers.
We present a Contrastive Attention-based Metric Learning approach to identify similar damage across character image pairs, which is sensitive to very subtle differences in glyph shapes, yet robust to various confounding sources of noise associated with digitized historical books. 
To overcome the scarce amount of supervised data, we design a random data synthesis procedure that aims to simulate bends, fractures, and inking variations induced by the early printing process.
Our method successfully improves downstream damaged type-imprint matching among printed works from this period, as validated by in-domain human experts. The results of our approach on two important philosophical works from the Early Modern period demonstrate potential to extend the extant historical research about the origins and content of these books. 
\end{abstract}

\section{Introduction}

\begin{figure}[t!]
\begin{center}
    \includegraphics[width=.47\textwidth]{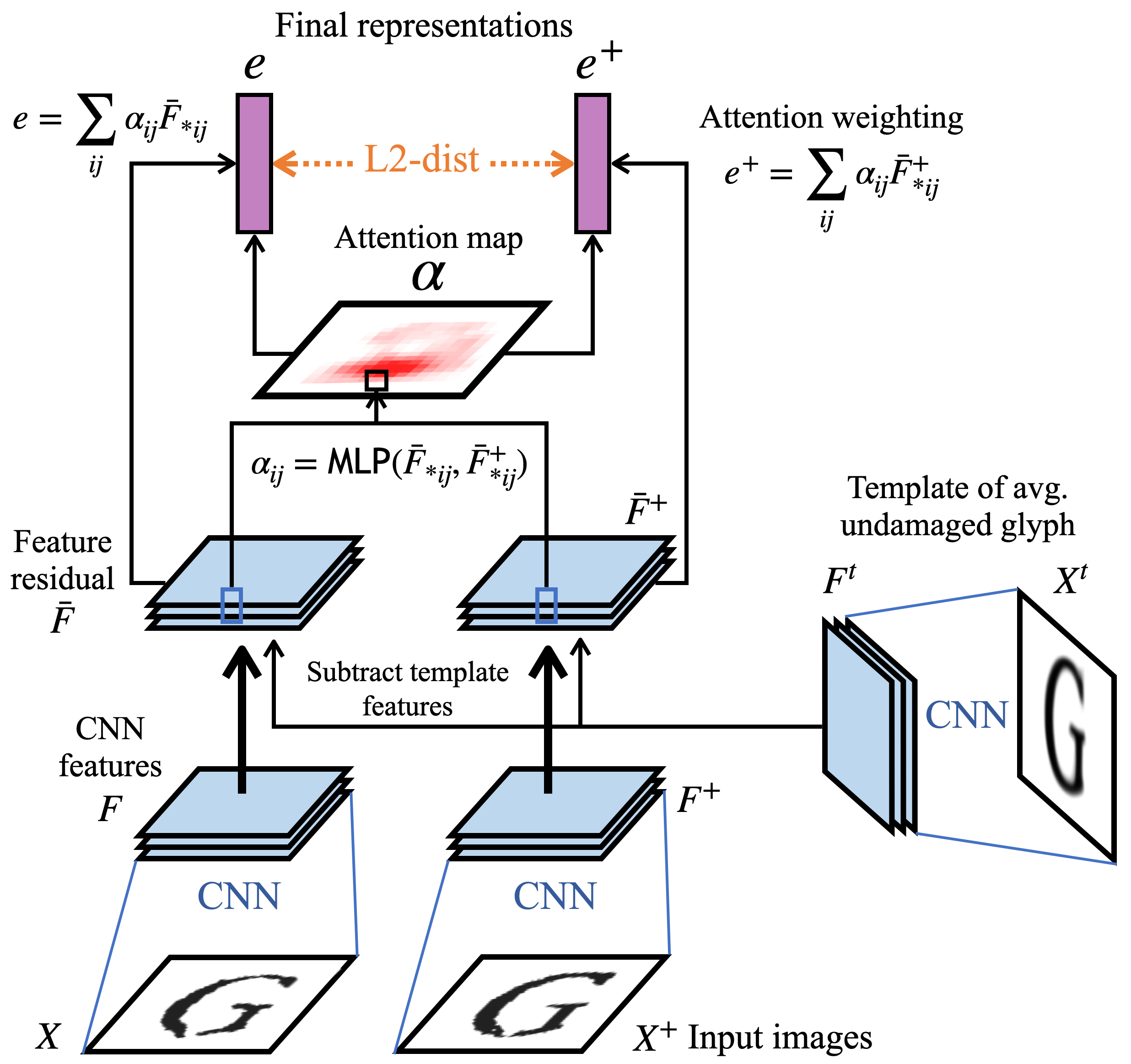}
    \caption{A depiction of the neural architecture behind Contrastive Attention-based Metric Learning (CAML), which we train to identify pairs of damaged type-imprints in early modern books that have similar damage and therefore likely originate from the same damaged piece of metallic type and printing apparatus. Our model uses a joint attention mechanism to attend to residual convolutional features after a template feature map has been subtracted from both inputs.}
    \label{fig:caml}
\end{center}
\end{figure}

\begin{figure}[t]
    \centering
    \includegraphics[width=0.47\textwidth]{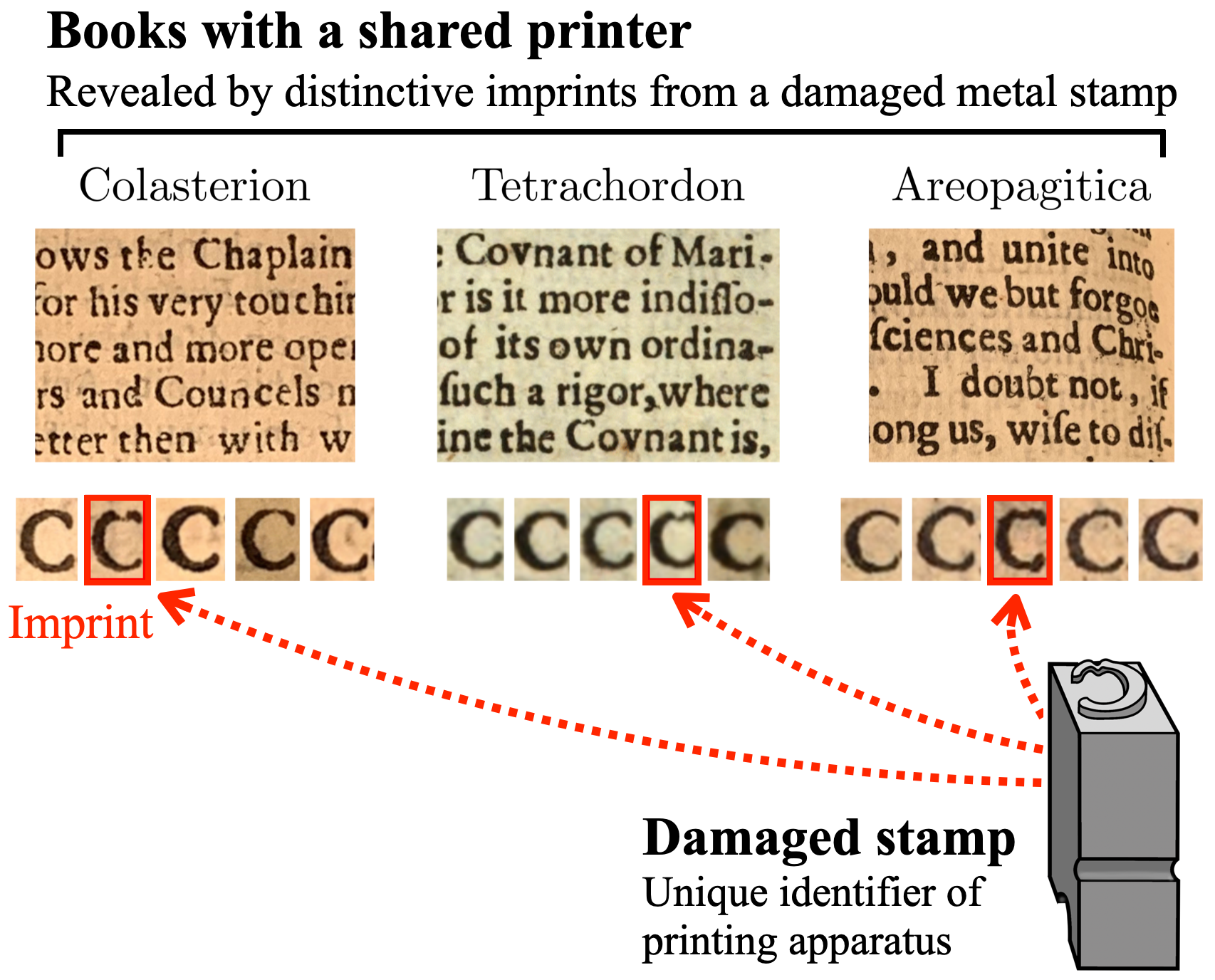}
    \caption{
    Although the three damaged type-imprints from \citet{warren2020damaged}'s study, in red, occur with different inking levels and imaging variation, the letters were pressed using the same type piece, evident from the corresponding damage at the top of the glyphs, suggesting that the respective books share the same printer. We are the first to approach this problem using machine learning, which could scale printer de-anonymization to tens of thousands of such books.
    }
    \label{fig:matches}
\end{figure}

A complete understanding of the content of surviving historical works requires knowledge of the context of their printing. 
This has long been acknowledged in the humanities, giving rise to \textit{analytical bibliography}, where physical evidence from printed books is used to understand their means of production, revealing incentives for their publication as well as cultural diffusion of language and knowledge. 
One of the main kinds of such physical evidence is character type-imprints pressed from damaged type pieces.
These anomalous type pieces, when linked across printed works, can reveal which works came from the same printing apparatus, as depicted in Figure~\ref{fig:matches}. %
Historians have, for example, leveraged the analysis of damaged type-imprints as a critical tool in identifying and studying clandestine printers who withheld their identities to avoid harsh penalties under censorship laws \citep{adams2010secret,como_radical_2018,warren2020damaged,warren2021canst}.

In this paper, we develop a computational model of the same analytic techniques used by bibliographers.
Our method \textit{automatically} compares type-imprint extractions across individual books, or larger collections of books, identifying pairs of \textit{damaged} type-imprints that match one another and therefore likely originate from the same, damaged piece of metallic type. %
Our approach enables \textit{computational} bibliographical analysis that has the potential to scale far beyond the limits of manual analyses -- for example, to the tens of thousands of censored early modern works that are of interest to social, cultural, and intellectual historians.

\begin{figure}[t]
    \centering

    \includegraphics[trim=16cm 14.2cm 16cm 9.2cm, clip,width=0.5\textwidth]{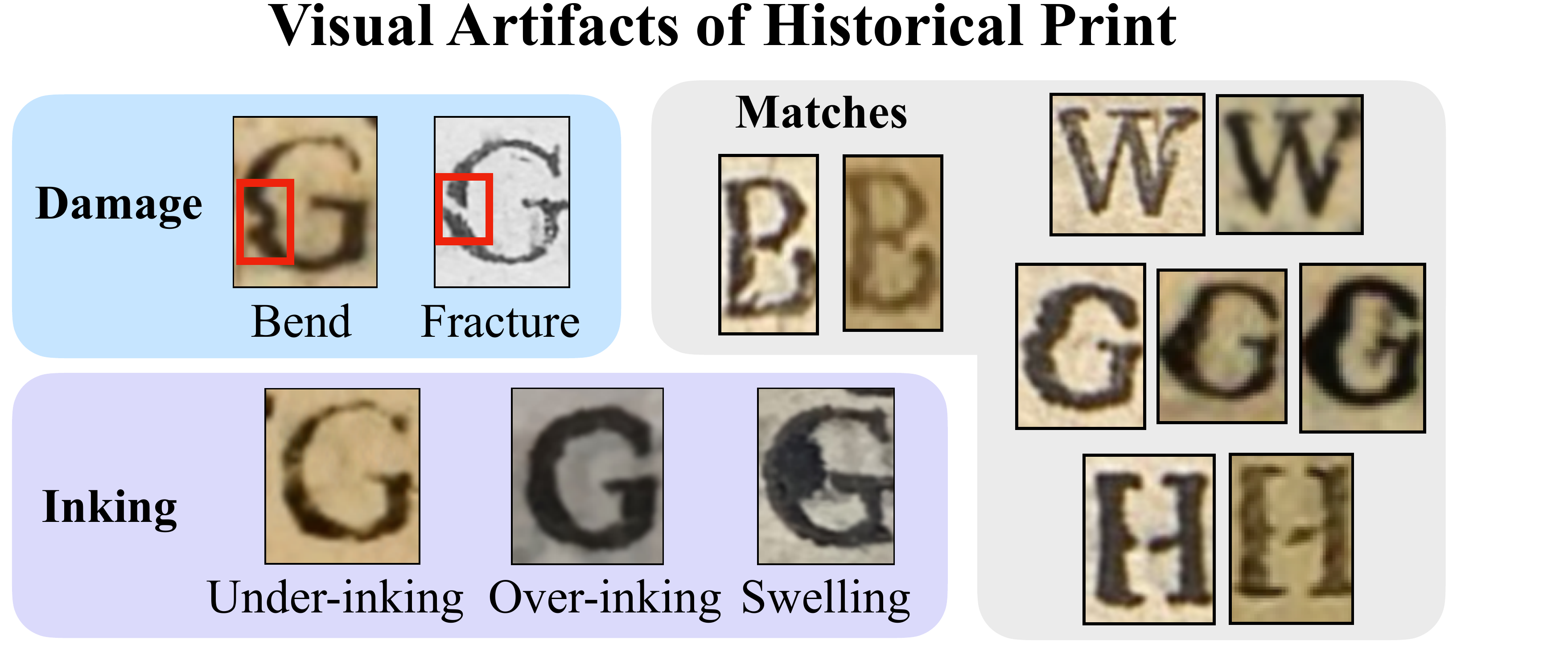}
    \caption{Bending and fracturing, the two main kinds of damage to type pieces in printed works, can be visually observed via the type-imprint left on a book's page. Three kinds of inking variation may complicate damage matching by either resembling or occluding damage. Under/over-inking generally appear evenly throughout a type-imprint, while swelling is more local. Example match groups are shown on right.}
    \label{fig:damage_and_inking}
\end{figure}

Matching damaged type-imprints across texts is challenging, since the variation in scale, position, font, and inking in scans of printed characters generally dominates the minute deviations due to localized damage that are of actual interest, as shown in Figs.~\ref{fig:damage_and_inking}~\&~\ref{fig:alignment}. 
We present a \textbf{Contrastive Attention-based Metric Learning} approach to identify similar damage across type-imprint pairs. Due to the dearth of training data consisting of matching damaged types, we design a specialized data generation process informed by human experts that aims to simulate bends, fractures, and inking variations induced by the early printing process. This lets us produce synthetic training data of image pairs with realistic matching damages. Then we develop a convolutional neural architecture that uses an attention mechanism for localized comparison of character pairs. By fitting this model to our synthetic data, we learn a metric appropriate for identifying similar, damaged type-imprints that generalizes to real data.
Our model is sensitive to very subtle glyph deformations, yet robust to various confounding sources of noise associated with digitized historical books. 
We evaluate our approach against other common methods for image comparison on a downstream damaged type-imprint matching dataset of English early modern (c.~1500--1800) books, a period when print censorship was prevalent. Our model's inferences are consistent with and support the findings of recent, manual bibliographical analyses of \emph{Areopagitica} \citep{warren2020damaged} and \emph{Leviathan Ornaments} \citep{warren2021canst}.
In a deployment case study, a broader set of type from \emph{Leviathan Ornaments} is matched against type from 138 candidate books and evaluated by an expert bibliographer.
Results suggest that our method can help scale printer attribution to the tens of thousands of anonymously printed early modern documents.

\section{Related Work}
Our work fits into a related line of work on computational analysis of historical documents under limited supervision---including, for example, unsupervised optical character recognition (OCR) techniques suited to early modern English historical documents such as \textit{Ocular}  \citep{berg-kirkpatrick-etal-2013-unsupervised,berg-kirkpatrick-klein-2014-improved}. 
\citet{garrette2015unsupervised} later proposed an OCR solution for code-switched historic texts using Ocular, followed by advances in improved orthographic transcriptions for OCR  \citep{garrette-alpert-abrams-2016-unsupervised}.
Separately, \citet{liu-smith-2020-detecting} investigated code-switching in historic German books with applications to OCR.
Neural techniques, such as the \textit{LatinBERT} masked language model for classical Latin language processing \citep{bamman2020latin}, and \textit{Lacuna} for low-resource document transcription for English and Arabic-script documents \citep{vogler-etal-2022-lacuna}, both utilize self-supervised pre-training to adapt to limited supervised resources.
In this work, we instead overcome data scarcity by synthesizing our data using a random augmentation procedure.

Recent work has also used machine learning to discover the origins of historical texts. 
\citet{assael2022restoring}'s \textit{Ithaca} uses deep neural networks for geographical and chronological attribution of ancient Greek inscriptions with the help of historians.
Closest in theme to our work is \citet{ryskina2017automatic}'s automatic compositor attribution, which considers the bibliography of the document by predicting the compositors of Shakespeare's \textit{First Folio} using orthographic and spacing features. 
While these approaches primarily focus on the text in the documents for computational analysis, we focus on often-overlooked but important complementary features related to the \textit{visual} appearance of books for computational attribution of print.

Our work also draws from a body of work on metric learning over images using contrastive approaches and data augmentation \citep{hadsell2006dimensionality, musgrave2020metric, weinberger2009distance, zhai2018classification} which are commonplace for many computer vision tasks like self-supervised learning of image representations \citep{chen2020simple}, image classification, and object localization \citep{ki2021contrastive}. While these approaches have been shown to be effective for capturing high-level features that characterize semantic variation among images, they are not suitable wholesale for our purpose which is to compare pairs of images based on subtle shape variations while remaining invariant to other dominant sources of variation. 

\section{Early Modern Print and Damaged Type Pieces}

We focus on a data domain that is highly relevant to the cultural analysis of censorship. From the advent of the printing press in the mid-1400s through the 18th century, printers declined to attach their names to tens of thousands, or about 25\%, of all known books and pamphlets. Reasons included controversial content, censorship laws, and piracy \citep{como_secret_2007, norbrook_areopagitica_1994, raymond_censorship_2017,woodfield_surreptitious_1991,towers_control_2003, mccabe_elizabethan_1981,bela_guiding_2016}.
As early as 1960, analytical bibliographers realized that distinctively damaged metal \textit{type pieces} used to produce the inked \textit{type-imprint} on a book's page could be recorded and compared across publications to uncover the hidden identities of printers \citep{mills_detective_1960}.
Essentially, once a metallic type piece is damaged, whether by being dropped on the ground, warped under pressure, or some other means, the type-imprints produced by the type piece become unique like a fingerprint as illustrated in Figure~\ref{fig:matches} and the top of Figure~\ref{fig:damage_and_inking}.
Manual forensic analyses of these most distinctive aberration patterns---bends and fractures, as well as other tell-tale cues identified by bibliographers, have produced evidence connecting anonymously printed works with known printers  \citep{weiss_shared_1992,van_den_berg_g._2004,achinstein_who_2013,garrett_how_2014,bricker_who_2016,lavin_printer_1972,adams2010secret,como_print_2012}.  
For example, Charlton Hinman's pioneering work in the 1960s exhaustively compared all letterforms across 55 copies of Shakespeare's First Folio using careful notetaking methods to uncover the collation process \citep{turner_reappearing_1966,hinman_printing_1963}.

Recent printer attribution work uses digital reproductions in addition to physical copies \citep{adams2010secret,como_radical_2018}. 
\citet{warren2020damaged, warren2021canst} go one step further, employing Ocular \citep{berg2013unsupervised} historical OCR for automatic extraction of type-imprint images, which they later \emph{manually} match to uncover the printers of important books printed in the early modern period. 

This type-imprint matching process remains the most labor-intensive part of the pipeline as every type-imprint image must be compared against an entire set of other type-imprint images.
In this work, we focus on developing automatic computational techniques that aim to scale the matching process by filtering the search space and suggesting candidates of matching damaged type-imprints across large collections of digitally archived books.

\begin{figure}[th]
    \centering
    \includegraphics[trim=36.2cm 8cm 32cm 6.5cm,clip,width=0.4\textwidth]{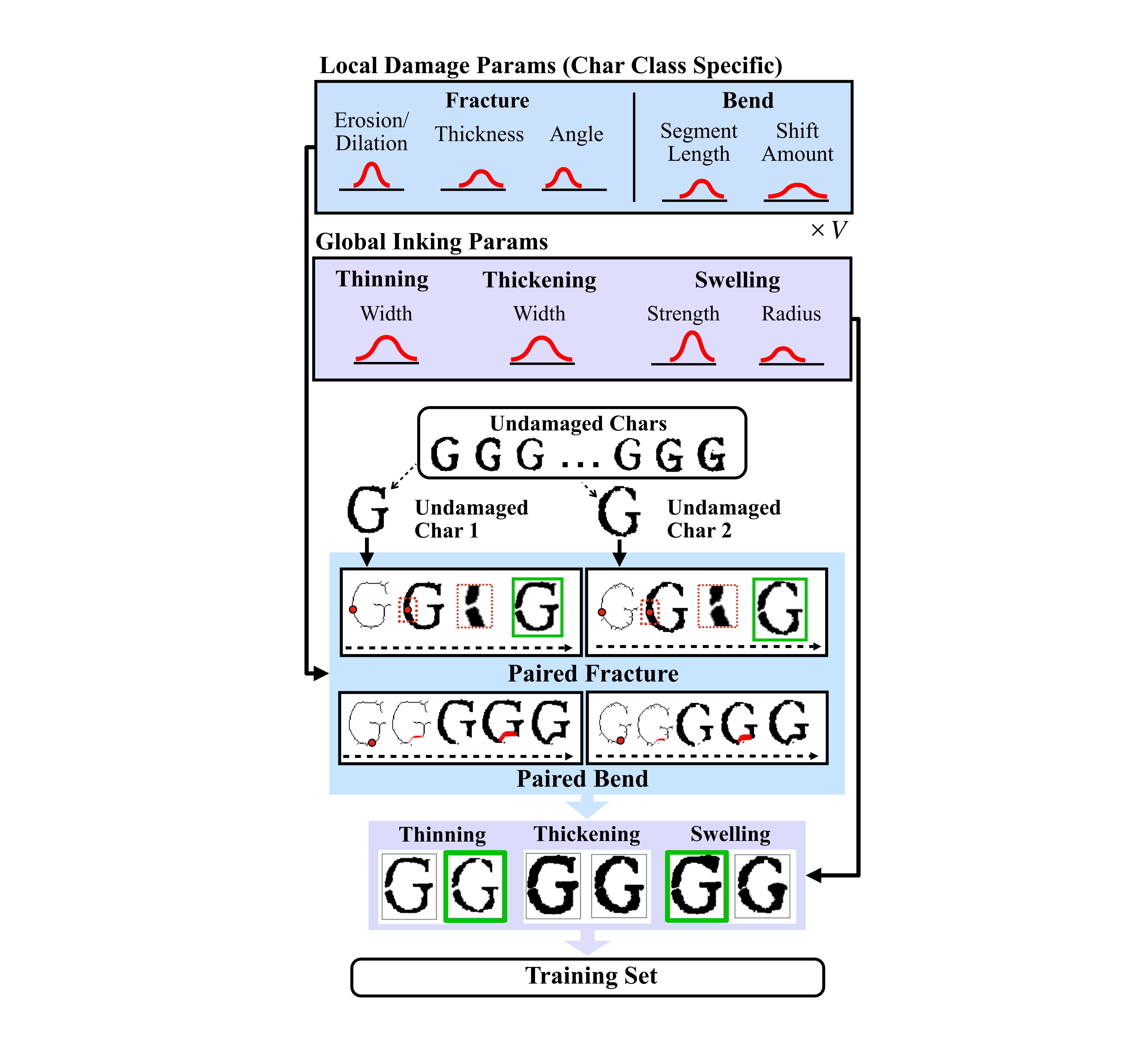}
    \caption{Our random data augmentation procedure models realistic bends, fractures, and inking to generate paired training data for learning to match damaged type-imprints. We sample two random undamaged images, perturb them with either a similar bend or a similar fracture (e.g., fracture here, green borders) using $V$ character classes' parameter sets, and apply one of three different types of random inking noise independently to each image (bottom, green borders).}
    \label{fig:perturber}
\end{figure}

\section{Our Computational Approach}  %

We use a neural model to learn a distance metric over pairs of type-imprint images. 
The metric learning framework facilitates treatment of matching as a search and rank problem, in which we embed a type-imprint and use it to \textit{query} and \textit{retrieve} the top-$k$ most similarly damaged type-imprints from the dataset.
Supervised training consists of learning to embed matched pairs comparably and unmatched pairs disparately using a margin loss.
However, this modeling approach faces two main hurdles: 
\begin{enumerate}
    \item \textbf{Spurious sources of deviations between type-imprints}, such as difference in scale, position, inking (Fig~\ref{fig:damage_and_inking}), font variation, character class due to imperfect historical typesetting and OCR errors, and digital imaging noise, mislead traditional metric learning approaches from isolating variation due to the shape of the underlying type pieces.
    \item The \textbf{lack of labeled data} complicates learning in the supervised metric learning framework. 
    Collecting high quality image scans of early modern books requires access to rare book libraries, identifying and annotating anomalous damage is labor-intensive, and manual pairwise comparison of every type-imprint is unfeasible at scale.

\end{enumerate}

In the following sections, we describe how we overcome these challenges.
First, we present our novel metric learning model (Figure~\ref{fig:caml}), which is able to overcome the confounding noise. 
Its restricted attention mechanism is well-suited to identifying shared local deviations in a pair of candidate images. 
Then, after describing the preparation of a collection of character type-imprint images, we describe how, by using domain-specific knowledge about the type piece damages, we generate realistic synthetic pairs of matching damaged type-imprints to train our model using a supervised objective.

\section{Contrastive Attention-Based Metric Learning}\label{sec:model}

As seen in the three matching character type-imprints in Figure~\ref{fig:matches}, identical type-imprints can be most easily recognized by the corresponding local deformations to character shape.
We use an attention mechanism to allow our model to jointly focus on \textit{corresponding locations} across a pair of images.
We introduce our Contrastive Attention-based Metric Learning (CAML) model, shown in Figure~\ref{fig:caml}, which computes a weighted attention map from the convolutional features of a type-imprint image pair to allow for more local comparison. 
At its core, our model takes as input a pair of type-imprint images we are interested in comparing and outputs a \emph{distance} between the pair. 
As shown in Figure~\ref{fig:caml}, $X$ is an image of a type-imprint of \texttt{G} with local damage at the bottom of the glyph and $X^+$ is another type-imprint image we would like to compare to. 
In this case, $X^+$ also exhibits similar local damage as $X$, so the desired output of the model, the L2-distance between the final image embeddings $e$ and $e^+$, should be small.

\paragraph{Template feature-map residuals} 
Most observable damage on a type-imprint image manifests as local deviations from an exemplary character shape.
In order to aid the model in identifying this local deviation from a standard, undamaged character shape, we additionally input a grayscale template image $X^t$ computed via the pixel-wise average over all training set images of the character.
We pass these three images through a deep convolutional neural network to get output feature representations $F, F^+$, and $F^t \in \mathbb{R}^{d \times h \times w}$ of the input images and the template image. 
Here $h$ and $w$ correspond to the height and width of the activations/feature-maps and $d$ is the depth of the feature-maps. 
We encourage the model to focus on deviation between input features and template features by computing \textit{residual feature-maps} $\bar{F}$ and $\bar{F}^+ \in \mathbb{R}^{d\times h\times w}$ via an element-wise difference between the input images' feature-maps and the template's feature-map: $\bar{F} := (F - F^t)$ and $\bar{F}^+ := (F^+ - F^t)$. 

\paragraph{Contrastive Attention Mechanism} 
Assuming that images $X, X^+, X^t$ have already been aligned (see Dataset section) so that the same pixel location in both images represents the same location on the type-imprint, we would like our model to be able to compare and contrast the same specific visuospatial features between residual feature-maps $\bar{F}$ and $\bar{F}^+$.
Our proposed contrastive attention mechanism, composed of a multi-layer perceptron (\textsc{mlp} in Fig.~\ref{fig:caml}) that computes an attention map $\hat{\alpha} \in \mathbb{R}^{h \times w}$ on the concatenated residual feature-maps $[\bar{F};\bar{F}^+]$, enables this comparison.
Each value $\hat{\alpha}_{ij}$ of $\hat{\alpha}$ represents a score for each spatial location, or ``feature pixel'', on the \textit{joint} location dimension of the residual feature-map:
$\hat{\alpha}_{ij} = \textsc{mlp}([\bar{F}_{*ij}; \bar{F}^+_{*ij}]).$

After normalizing these attention scores, we obtain a weighed attention map $\alpha \in \mathbb{R}^{h \times w}$ over such feature pixels. 
We compute the final $d$-dimensional output image embeddings $e$ and $e^+$ by attending to $\bar{F}$ and $\bar{F}^+$ respectively: $e := \sum_{ij} \alpha_{ij} \bar{F}_{*ij}$, and $e^+ := \sum_{ij} \alpha_{ij} \bar{F}^+_{*ij}$. 
Finally, the model returns the Euclidean distance between the embeddings $\|e - e^+\|_2$, representing the distance between the type-imprint image pair.

\paragraph{Training CAML}
We train CAML with the popular triplet loss \citep{weinberger2009distance}, which operates on an anchor/query embedding $e$ along with the embedding $e^+$ of a candidate image that matches the anchor and a non-matching candidate image's embedding $e^-$. This results in the following loss: $\max\big(\|e - e^-\|_2 - \|e - e^+\|_2 + m, 0\big)$, which focuses on minimizing the Euclidean distance between the anchor and the positive matching images' embeddings, and maximizing the distance between the anchor and the non-matching images' embeddings, such that the positive and negative examples are separated by a margin of at least $m$.
We sample negative examples uniformly at random from our batch. %

\section{Synthesizing Damaged Type-Imprint Pairs}

Given pairs of known matches and non-matches, we can train CAML. 
However, annotated data of this form is extremely sparse and difficult to collect.
In this section, we describe how we first create a small collection of \textit{unpaired} damaged character type-imprint images from early modern books of both known and unknown printer origins.
Then we describe a random automatic process for synthesizing \textit{paired} supervised match data from this collection for training CAML.

\subsection{Extracting Type-Imprint Images from  Books} \label{sec:dataset}

We obtain page image scans from 38 different English books printed from the 1650s--1690s by both known and unknown printers of historical interest. 
The materials were printed mainly in London after its  civil war when the explosion of news, pamphlets, and cheap print are said to have led to England's `reading revolution'  \citep{como_radical_2018, sharpe_reading_2000, achinstein_milton_1994, raymond_pamphlets_2003, zaret_origins_2000}.
In contrast to resources like Early English Books Online (EEBO), used in \citet{mak_archaeology_2014}, %
our images are higher resolution to facilitate fine-grained comparison of anomalous type-imprints.

Next, we use the Ocular OCR system \citep{berg-kirkpatrick-etal-2013-unsupervised} to extract segmented character images from the color page photographs and select a subset of 16 capital letters, which tend to exhibit the most recognizable damaged compared to their relatively infrequent occurrence in printed English. 
Then, we align the resulting character images using learned rotation, offset, size, and scale random variables from \citet{goyal2020probabilistic}'s recently proposed generative font clustering model for typographical analysis of early modern printing, which significantly reduces the variance in size, offset, skewness, and rotation.

\subsection{Supervised Data via Paired Damage Synthesis} \label{sec:perturber}

In order to train our matching system, we require a dataset consisting of pairs of matching damaged characters --- we propose a technique for synthesizing such pairs. 
Inspired by work in learning disentangled representations of morphologically perturbed MNIST images \citep{castro2019morphomnist} and synthetically shifted images of characters from historical books for font discovery \citep{goyal2020probabilistic}, we design a detailed random process for perturbing character images with \textit{paired} realistic damage and inking variation observed in printing press era books.
The generating process is depicted in Fig.~\ref{fig:perturber} and described in this section.

First, we sample two random, undamaged letter type-imprint images from different books from the annotated collection described above (as depicted at the top of Fig.~\ref{fig:perturber}).
By sampling the type-imprint images from different books instead of using the same book or even the same, duplicated character type-imprint image,\footnote{In fact, training on paired data synthesized from the same character type-imprint image does not generalize well to real datasets.} we force our model to discriminate between the synthesized local damage patterns instead of specious features arising from the book scanning process itself. 
We take the union of the sampled type-imprint pair and extract its skeleton (henceforth referred to as \textit{union skeleton}), which makes it easier to produce a bend/fracture on the same location of the pair.\footnote{We use \textit{scikit-image} morphology \citep{scikit-image}.}
We randomly apply one of two kinds of damage--bends or fractures--followed by inking noise, as described below.

\paragraph{Bends}
In the case of a bend (Fig.~\ref{fig:damage_and_inking}, top left), we first sample the \textit{length of the segment} to be bent and the \textit{amount to shift} the segment (both as relative percentages of skeleton height) from the respective character class distributions (Fig.~\ref{fig:perturber}, right). 
Next, we iteratively sample for the best set of midpoints and endpoint locations on the \textit{union skeleton} and map them back to the original images using the nearest neighbor on each skeleton.
We morph the sampled \textit{segment length} by drawing a Bezier curve between the midpoint and a shifted midpoint computed with the sampled \textit{shift amount}.
We increase the bent skeleton's thickness until it reaches the unbent character's thickness, as measured by a Euclidean distance transform.

\paragraph{Fractures}
We sample a center point on the \textit{union skeleton} and map it back to the original images in a similar fashion as in the bends. 
For each type-imprint in the pair, we locally erode the image by a sampled \textit{percentage of its measured mean character thickness} in a small window around the center point on the character skeleton.
Then, we sample a random \textit{angle} and draw a circular brush stroke through the center point at this angle, similar to \citet{castro2019morphomnist}, with sampled \textit{thickness} proportional to mean character thickness.
Finally, the image is locally dilated back to its original thickness amount. %
By placing the brush stroke between local erosion/dilation, the sharp edges introduced by the circular brush stroke are removed and the damage looks less artificial.

\paragraph{Inking}
In early printing press era books, natural variation in a type piece's inking level can cause significant visual differences in type-imprints produced from the same underlying metal type piece, as shown in Fig.~\ref{fig:damage_and_inking}.
Under-inking can create superficially similar effects as damage, whereas over-inking can occlude actual damage to the metal type piece. Either the whole image could be over or under-inked or local sections of glyphs could exhibit inking variations.
We over- or under-ink both undamaged and artificially damaged character type-imprint images \textit{independently} by dilating or eroding the image by a sampled percentage \textit{amount} of the type-imprint's mean thickness, which evenly thickens or thins the character (Fig~\ref{fig:perturber}, left).
Similar to \citet{castro2019morphomnist}, for local inking, we perturb a character type-imprint image 
by first uniformly sampling a random location on the character skeleton, sampling a \textit{strength} magnitude to increase inking, and \textit{radius} amount, followed by warping of image coordinates around the sampled location.
Modeling these inking variations in our perturbation generating process exposes our learned matching models to this deceptive real-world noise.

\paragraph{Setting Parameters}
Instead of learning these damage and inking parameters, which would be difficult given the limited  labeled data, we consult humanities scholars with expert domain knowledge.
We set parameters of the truncated normal distributions controlling the image operations (shown in Fig.~\ref{fig:perturber}) through multiple iterations of tuning, sampling, and evaluation.
In each round, we display comparisons between hundreds of samples of real and synthesized damages and have annotators evaluate the fidelity of the generated samples.

\section{Experimental Setup}\label{setup}

In this section, we first describe the real-world datasets containing a small amount of manually identified matches and damaged types in previous bibliographic studies that we use to evaluate our approach on downstream matching of type-imprint images. We use these datasets to construct various scenarios, ranging from highly optimistic to more realistic settings, pertaining to the quality of the candidate sets we query against for our matching experiments. 
Then, we present strong baselines that are prevalent for computational image comparison for empirical comparison against our proposed approach.

\subsection{Ground Truth Evaluation Datasets}
We use two different hand-curated datasets from recent bibliographical studies that manually identified and matched damaged type-imprints for attribution of two major early modern printed works \citep{warren2020damaged,warren2021canst}. 
Per-character dataset statistics are presented in  Table~\ref{tab:test_set_stats_per_char}.

\paragraph{{\areo} validation set}
We collect a small \emph{validation} set of the manually identified type-imprint matches used in the study for printer attribution of John Milton's anonymously printed {\areo} \cite{warren2020damaged}.
Specifically, we focus on four uppercase characters \texttt{D}, \texttt{F}, \texttt{G}, and \texttt{M} whose damaged type-imprints were compared across books from known printers of interest for this study.
Overall, this dataset contains 128 total match groups with 159 pairwise queries.

\paragraph{{\lev} test set}
We construct our \emph{test} set from the expert-curated set of matches manually identified in a recent bibliographical study, in which \citet{warren2021canst} established that the book was printed in a single print shop by John Richardson in 1695--1696, thus refuting the attribution of \citet{malcolm_making_2008} to a different printer, John Darby in earlier work by amassing evidence from damaged type-imprints matched manually within {\lev} and across other books by known printers from the suspected time period of its printing.
Overall, this dataset contains 217 total match groups with 858 pairwise queries.

For the purpose of damaged type-imprint retrieval, we treat the known damaged type-imprints in {\areo} and {\lev} as queries against which we attempt to retrieve the matching damaged type-imprints from candidate sets of interest. 
For the {\areo} validation set, we simply match against the other \textit{queries}---i.e. every candidate is a match for at least one query. Note, this setup is unrealistically easy, and as we will see in experiments, simple baselines do unrealistically well. However, we find it useful to leverage {\areo} for validation and early stopping. 
{\lev} represents our main test evaluation set. Here, we conduct experiments with two different setups that include different types of realistic candidates sets: (1) {\bf Strong negative} (Lev-Strong), in which the candidate set is composed of both ground truth positive matches and 1000 random type-imprint images from other books printed by Darby and Redmayne, who were both suspected of being the printer behind {\lev} until \citet{warren2021canst}'s study, and (2) {\bf Mix negative} (Lev-Mix), where the candidate set consists of both ground truth positive matches, 500 random images from Lev-Strong, and 500 random images from books printed by Robert Everingham, who used type pieces in a different font and was never considered to be involved in the printing of {\lev}. These test setups are more realistic because they contain a larger variety of negative candidates and are based on the actual data bibliographers have combed through to identify matches. The latter setting, Lev-Mix, is possibly the most difficult because it includes the most variability in negative examples, making spurious matches using simple methods more likely. 

\subsection{Baseline Models}
In order to show the effectiveness of our approach on matching subtle damages between type-imprints in the presence of multiple confounding sources of noise, we compare it to other prevalent approaches for image comparison.

\paragraph{Image L2 Distance (L2)} 
Instead of learning a metric over the set of training images, this method compares the aligned (see Dataset section) query images to the candidate images by  computing the Euclidean distance between the two images and outputs a ranked list of candidates for each query image.

\paragraph{Embedding Triplet Loss (Emb)} In a large metastudy on metric learning for images, \citet{musgrave2020metric} report that classic embedding based triplet losses are highly effective and closely match the state-of-the-art on metric learning tasks. 
Therefore, we compare our method to an \emph{embedding based triplet loss} approach as described in \citet{weinberger2009distance}. 
This approach is trained with the same loss as CAML and is similar to the approach described in the CAML section, but the images (anchor, match, non-match) for triplet loss training are processed independently of each other by a CNN with no attention mechanism to produce their respective embeddings.

For fair comparison, the architecture, artificial data creation and negative mining strategies are the same as our approach. 
The one \emph{major difference} between our approach (CAML) and this baseline approach is that our model considers both the anchor and the candidate jointly via the described attention mechanism to yield the relevant embeddings whereas this baseline learns the embeddings for each triplet image independently.
As observed in the results, this difference makes our model more suitable for metric learning over character images with very subtle local variation.

\paragraph{Stacked BCE} To investigate the effect that our attention-based architecture design and contrastive loss function has on performance, we train the same convolutional neural network feature extractor to classify positive and negative pairs \citep{zhai2018classification} of images with a binary cross-entropy loss. Instead of encoding the 1-channel images separately, we stack the images on separate channels before inputting to the CNN.

\section{Quantitative Results}
In this section, we describe our quantitative results on the ground truth evaluation data presented in Experimental Setup. 
We report all-pairs \emph{Recall@k=5} micro-averaged over the character classes in each dataset configuration in our evaluations so every damaged type-imprint serves as the query image once.
We train each model for 60 epochs and early stop using the best {\areo} validation set recall.\footnote{Code located at \url{https://github.com/nvog/damaged-type}.}

In Table~\ref{tab:main_results}, we compare results using the L2 baseline on the aligned images, Stacked BCE classifier, Triplet Embedding model, which is equivalent to CAML without the spatial attention mechanism, and our proposed CAML method. 
First, we observe that CAML outperforms the nearest baseline approach by at least 15--18 points on recall @ 5 on the {\lev} test settings. 
The next closest methods for this setting are the L2 baseline and the Emb model, which both tend to lack the ability to focus on local damage similarities.
We hypothesize that character positioning and inking dominates most of the distance for these methods.
This explanation is plausible when considering the poor synthetic validation performance, which is a dataset consisting of type-imprints from different underlying fonts to force models to focus on local similarities in lieu of font shape.
Also, this is supported by the high recall on the {\areo} validation set---all of the characters in it are hand-identified ground truth matches meant to be as similar as possible to make a convincing bibliographical argument \citep{warren2020damaged} and no other misleading candidate images need to be filtered out.
The Stacked BCE model, while capable of driving its classification loss near zero on the training set, has a lot of difficulty ranking unseen data and performs  worst of all methods.

\begin{table}[t]
    \small
    \centering
    \begin{tabular}{l | c c c c}
    \toprule
    Method &  Syn Valid & Areo & Lev-Strong & Lev-Mix \\
    \midrule
    L2 Aligned   & 5.53 & \bf 43.21 & 39.91 & 40.99 \\
    Stacked BCE & 4.62 & 1.78 & 1.10 & 0.96  \\
    Emb  &  5.72 & 40.76 & 33.53 & 33.00 \\
    \textbf{CAML} & \bf 35.50 & 39.42 & \bf 58.15 & \bf 56.08 \\
    \bottomrule
    \end{tabular}
    \caption{Recall @ 5 results micro-averaged over 4 character classes for {\areo} ground truth validation set, and 16 character classes for both the synthetic validation set and {\lev} ground truth test set. Both the Emb and CAML models use convolutional residual features.}
    \label{tab:main_results}
\end{table}

\begin{table}[t]
    \small
    \centering
    \begin{tabular}{l | c c c c }
    \toprule
     Residual & Syn Valid & Areo & Lev-Strong & Lev-Mix \\
    \midrule
    None & \bf 50.62 & 33.63 & 62.88 & 61.78 \\
    Input & 47.62 & 38.75 & 52.33 & 49.18 \\
    CNN & 45.00 & \bf 43.88 & \bf 67.81 & \bf64.79 \\
    \bottomrule
    \end{tabular}
    \caption{Micro-averaged recall @ 5 independent ablation study comparing different residual methods in the CAML model. Results on \texttt{D}, \texttt{F}, \texttt{G}, \texttt{M} character subset.}
    \label{tab:ablate_residual}
\end{table}

\paragraph{Ablation} %
We perform two kinds of ablative studies on our model. The first study (Table~\ref{tab:ablate_residual}) concerns with how we obtain the residuals of the image representations through subtracting the template character features. We compare 3 settings: (1) no template residual, (2) an input residual where the template image pixels are subtracted from each input image's pixels, and (3) CNN feature residual (our final model), where the learned template image's CNN features are subtracted from the learned input image's CNN features before performing spatial attention. We observe that providing information about template shape of the character class is important for our model's success as shown by the superior performance of the \emph{CNN residual} variant. Interestingly, not using any residuals performs better than subtracting the templates from the input image, suggesting that the \emph{input residual} variant suffers from information loss before CNN feature computation---perhaps due to misaligned characters or inking variation.

Additionally, we perform a small ablation analysis to study training CAML without global inking synthesis such as Thinning and Thickening from Fig.~\ref{fig:perturber}. Compared to our best model which uses this augmentation, most of the decrease in performance is on synthetic validation (3 points R@5) and the Lev-Mix setting (0.68 R@5). 

\begin{figure}[h]
    \centering
    \includegraphics[trim=54.5cm 118.3cm 44cm 1.55cm,clip,width=\linewidth]{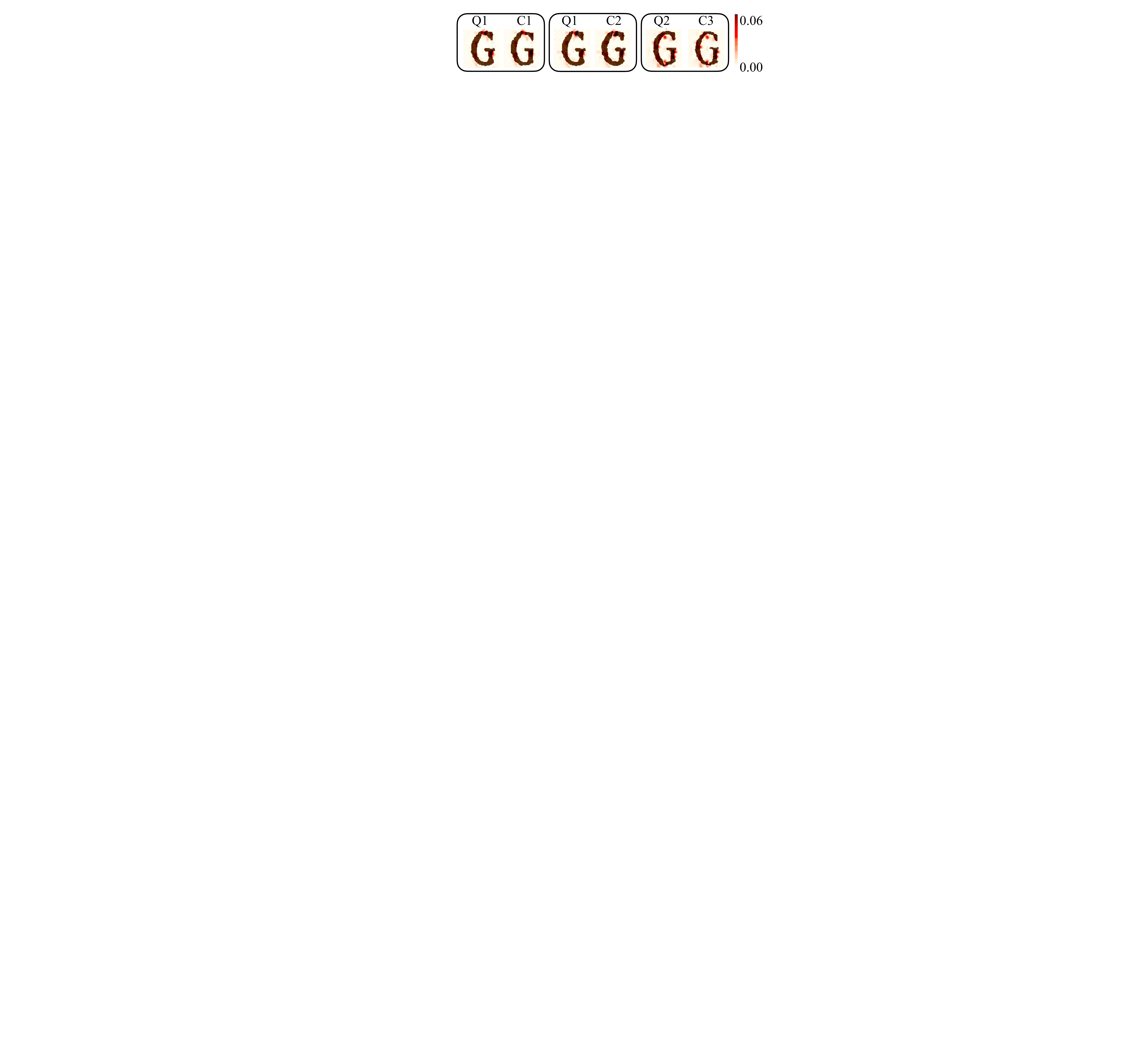}
    \caption{CAML attention for 3 query/candidate pairs.}\label{fig:attention_viz}
\end{figure}
\paragraph{Attention Visualization} In Figure~\ref{fig:attention_viz}, we visualize the spatial attention weights on a few top-ranked \texttt{G} type-imprint images from the Lev-Mix test set.
We compare two pairs (left, middle) using the same `Query 1', while the third pair (right) uses a different `Query 2'. The highest attention weights are clearly located in locations with noticeable damage, along with other areas that help the model differentiate glyph shapes. Weights appear mostly similar for the pairs with the same query, but adapt drastically for the other pair.

\section{Deployment Study: CAML for Attribution}

\begin{table}[t]
    \small
    \centering
    \begin{tabular}{l c c c c c c}
    \toprule
     \multirow{2}{*}{Printer} & \multicolumn{2}{c}{Top-1} & \multicolumn{2}{c}{Top-5} & \multicolumn{2}{c}{Top-10} \\
      & Lo & Hi & Lo & Hi & Lo & Hi \\
    \midrule
     Richardson & 7 & \textbf{11} & 17 & \textbf{23} & 40 & \textbf{31} \\
     Richardson \& Holt & 0 & 0 & 2 & 0 & 2 & 0 \\
     Darby & 13 & 3 & 21 & 7 & 28 & 9 \\
     Redmayne & 0 & 0 & 0 & 0 & 0 & 0\\
     Everingham & 0 & 0 & 5 & 1 & 9 & 1 \\
    \bottomrule
    \end{tabular}
    \caption{CAML deployment findings aggregated across matches for known printers on {\lev}. Expert is asked to annotate both low (lo) and high (hi) confidence matches returned by system. The most high confident matches are bolded, confirming \citet{warren2021canst}'s manual  attribution to the printer John Richardson.}
    \label{tab:deploy}
\end{table}

For analyzing how our model can aid bibliographical research, we design a deployment and expert evaluation case study to confirm the manual printer attribution of {\lev} \citep{warren2021canst}.
Using an unannotated set of 138 books scanned by various libraries and printed by 4 suspected printers of {\lev} and 1 unrelated printer, we aim to simulate the entire process of human-in-the-loop printer attribution, from query selection to matching.
Until now, such a process has required much manual effort.

\paragraph{Query Selection and Matching} First, we start with query selection, which involves choosing hundreds of type-imprints exhibiting anomalous bends and fractures in {\lev}, the anonymously printed book of interest.
Instead of manually identifying these by sorting through tens of thousands of characters, we train a CNN-based classifier with the same architecture as our matching models on thousands of unpaired, labeled images from other 17th-century books to rank all {\lev} type-imprints by damage intensity.
An expert bibliographer then selects a few hundred of the top results to create a query set of 246 images.
Next, using the L2 distance on CAML's image embeddings, we generate the top-10 ranked candidates for each query image against a set of 518,891 total type-imprint images.

\paragraph{Annotation Findings} We setup an annotation interface and ask a bibliographical expert to  annotate matching type-imprints among the retrieved candidates as either not a match, or as a \textit{high} or \textit{low} confidence match without access to book/printer name or scoring information (see Appendix for Fig.~\ref{fig:annotation} and annotator feedback).
In Table~\ref{tab:deploy}, we present aggregated match counts by summing up the counts for each printers' books.
Of the 246 unique damaged type-imprint queries, at least one match was found in 80 of them (32.5\%).
 Among the suspected candidate printers, John Richardson has the largest amount of high confidence matches. 
In contrast to John Darby, who \citet{malcolm_making_2008} attributed the book to, John Richardson has 22 more high confidence matches in the top-10.
While John Redmayne printed another clandestine version of Hobbes' \emph{Leviathan}, zero matches surfaced in our investigation.
Robert Everingham's books, which tend to use different sets of fonts entirely, only surfaced a single high match.
Ultimately, evidence strongly suggests that John Richardson did, in fact, print {\lev}.

\section{Conclusion}
We demonstrate that machine learning can be successfully applied to printer attribution by learning to match damaged type-imprint images in early modern books.
By attributing early modern print at scale, we can begin to uncover the hidden figures behind the tens of thousands of clandestinely printed works from the period, which amounts to roughly 25\% of documents.
Ultimately, this translates into more opportunities to discover significant historical networks of early print media involving printers, authors, and arguments.

\section*{Acknowledgments}
This project is funded in part by the NSF under grant 1936155, and by the NEH under grant HAA-284882-22. We would like to thank Ciaran Evans, DJ Schuldt, Kari Thomas, Laura DeLuca, and the rest of the Print \& Probability team, along with Berg Lab, and the anonymous reviewers for their assistance and helpful feedback.

\bibliography{biblio}
\clearpage
\appendix

\begin{table}[t]
    \small
    \centering
    \begin{tabular}{c c c c}
    \toprule
       Title                  & Char &  Match Groups & Pairwise Queries  \\
    \midrule
       \multirow{4}{*}{\areo}  &  \texttt{D} & 38 & 57 \\
                                              & \texttt{F} & 33 & 47 \\
                                               & \texttt{G} & 40 & 44  \\
                                              & \texttt{M} & 17 & 11 \\ \cmidrule{2-4}
        & Totals: & 128 & 159 \\
    \midrule
        & \texttt{A} & 30  & 134   \\
        & \texttt{B} &  14 & 46   \\
        & \texttt{C} &  22 & 99   \\
        & \texttt{D} &  14 & 59   \\
        & \texttt{E} &  4 & 19  \\
        & \texttt{F} &  15 & 53   \\
        & \texttt{G} &  17 & 59   \\
       \emph{Leviathan} & \texttt{H} &  8 & 19  \\
       \emph{Ornaments} & \texttt{K} & 5  & 18   \\
        & \texttt{L} &  17 & 66   \\
        & \texttt{M} &  13 & 47   \\
        & \texttt{N} &  7 & 20   \\
        & \texttt{P} &  13 & 75   \\
        & \texttt{R} &  13 & 47   \\
        & \texttt{T} &  14 & 58   \\
        & \texttt{W} &  11 & 39   \\ \cmidrule{2-4}
        & Totals: & 217 & 858 \\
    \bottomrule
    \end{tabular}
    \caption{Per-character ground truth dataset statistics of {\lev}, which is the main test set used for quantitative evaluation. Aggregates are reported at bottom.}
    \label{tab:test_set_stats_per_char}
\end{table}

\begin{figure}[t]
    \centering
    \includegraphics[width=.43\textwidth]{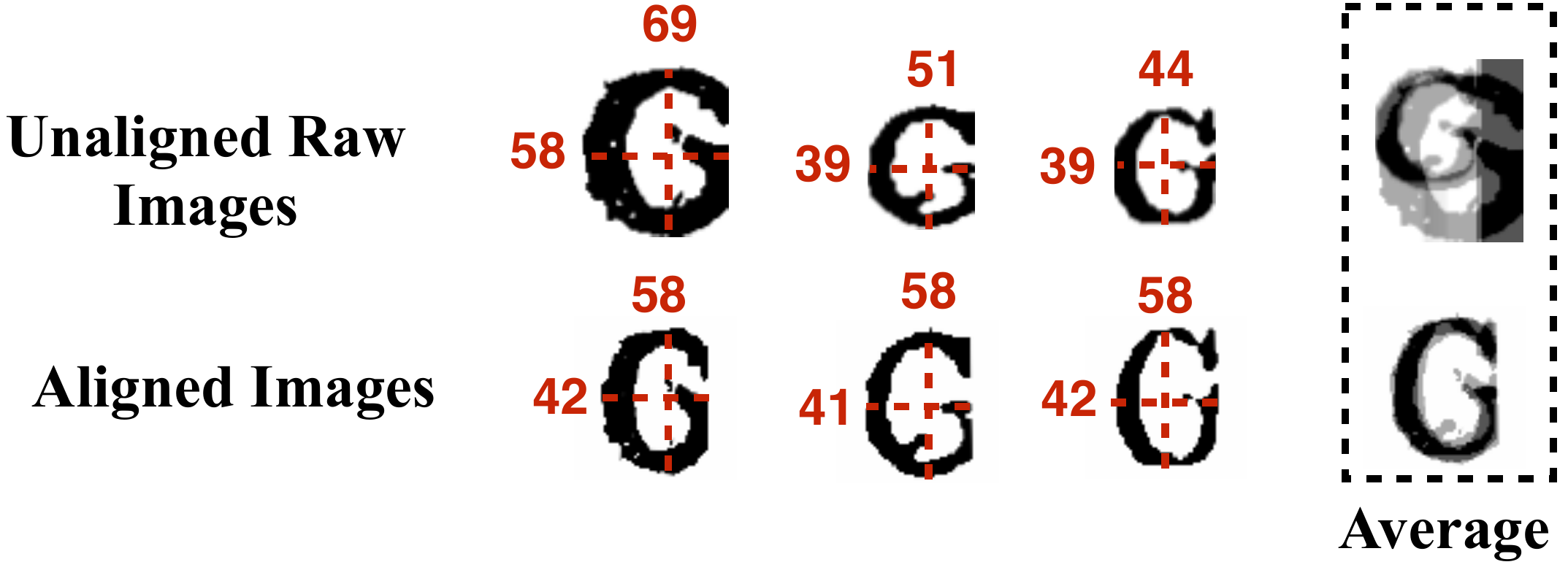}
    \caption{Effects of aligning character type piece type-imprints standardizes their height/width, offsets, shear, \& rotation to a common size/orientation for more consistent glyph comparison across models \citep{goyal2020probabilistic}.}
    \label{fig:alignment}
\end{figure}

\section{Training Details}
All models are built using a number of CNN `blocks' containing 3 convolutional layers, separated by batch normalization \citep{ioffe2015batch} and ReLU non-linearities.
Convolutional layers in the block all contain 128 filters each, kernel sizes of 3, strides of 1, and padding of 1, such that the only spatial dimension reduction is from the max pooling layers with kernel size and stride size of 2 with no padding.
For Emb model, we use 4 blocks followed by a flatten operation and linear layer with 2048-dimensional input.
Emb models use 128-dimensional image embeddings.
For CAML models, we use only 2 blocks in order to avoid reducing the spatial resolution of the feature maps before the attention operation.
CAML's Attention module is a 4-layer MLP with hidden sizes of 256, 128, and 64 units separated by Tanh activations.
We use a batch size of 64, Adam \citep{kingma2014adam} with a learning rate of 0.0001 and grid search over number of blocks $\{2, 3, 4, 5\}$, number of convolutional layers per block $\{2, 3, 4, 5, 6\}$, CAML's attention softmax temperatures $\{0.1, 0.5, \textbf{1.0}\}$, and triplet loss margins $\{0.1, 0.2, 0.3, 0.4, 0.5\}$.
We find a margin of 0.3 to yield consistently good performance across models.
All hyperparameters are tuned on the {\areo} validation set.

\section{Annotator Feedback}
\begin{figure*}[t]
    \centering
    \includegraphics[width=0.98\textwidth]{"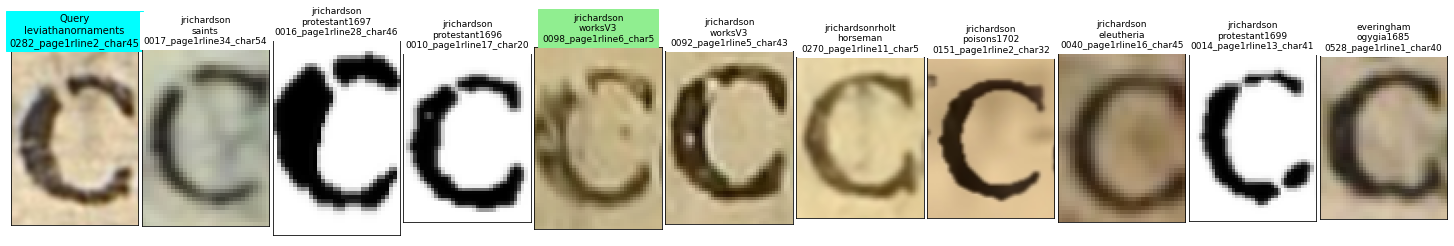"}
    \includegraphics[width=0.98\textwidth]{"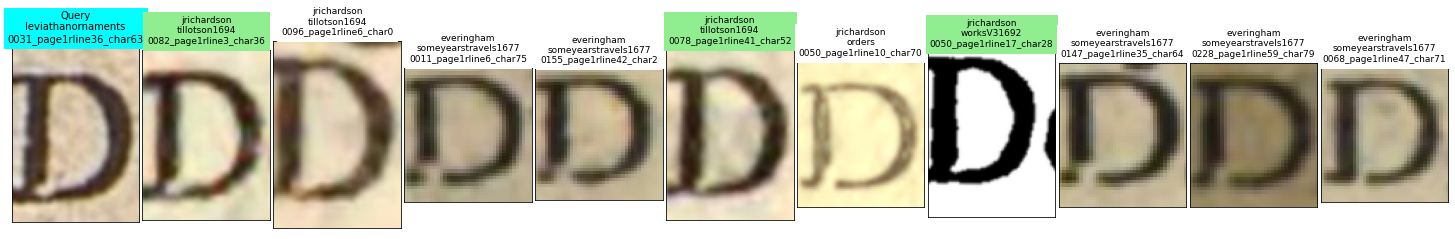"}
    \caption{Selected results from annotation of our deployed system, in which an expert bibliographer must decide whether the query character type-imprint image (in cyan, far left) matches any of the top-10 ranked candidate character type-imprint images as scored by our model (right of the query). Matches annotated as `high confidence' are shown in green.}
    \label{fig:annotation}
\end{figure*}

In Figure~\ref{fig:annotation}, we show two rows from the expert annotation task, which underscore how difficult type-imprint damage identification can be. 
For example, all 5 of the first 5 \texttt{C} candidates include a break in the upper stroke indicating high plausibility of matching the query image.
Yet the bounding boxes for the first two make it difficult to assess the \texttt{C}s' terminals for comparison.
The resolution on the first image is poor and the binarization in the next two makes high confidence difficult. 
The fifth one could very well be a match, yet the ink doesn't break as cleanly as it does in the query image so it was not selected.

On the second row of \texttt{D}s, the query contains a counter (white space) that extends into the juncture of the stem, bowl, and lower serif of the character.
In this case, all of the results contain such damage.
However, the selected high confidence matches appear thinner than the others.
The one exception is the second option from the Tillotson book, which could very well be the same piece of type, but either the inking or some other feature means the counter didn't extend into the lower bowl quite as prominently.

\end{document}